\documentclass[letterpaper, 10 pt, journal, twoside]{IEEEtran}

\usepackage{etex} 

\usepackage{lipsum}
\usepackage[utf8]{inputenc}
\usepackage{makeidx}
\usepackage{amsmath,bm}
\usepackage{amssymb}
\usepackage{color}
\usepackage{ifpdf}
\usepackage{wrapfig}
\usepackage{graphicx}
\usepackage{sidecap}
\usepackage[]{threeparttable}
\usepackage{multirow}
\usepackage{relsize}
\usepackage{algorithm}
\usepackage{algorithmic}

\usepackage{pgfplots}
\usepackage{mathrsfs}
\usepackage{bbm}
\usepackage{mathtools}
\usepackage{upgreek}
\usepackage{trfsigns}
\usepackage{arcs}
\usepackage{pgf}
\usepackage{lscape}
\usepackage{array, makecell} %
\usepackage{todonotes}
\usepackage{caption}
\usepackage{subcaption}
\usepackage{svg}
\usepackage{cite}
\usepackage{physics}
\usepackage{hyperref}
\usepackage{footmisc}
\usepackage{xspace}
\usepackage{xparse}
\usepackage{fancyvrb}

 \usepackage{caption}
\newcommand{\comment}[1]{}

\NewDocumentCommand{\ShowInline}{v}{%
#1%
}
\IEEEoverridecommandlockouts                              
\usepackage{siunitx}                        
\usepackage{pythonhighlight}
\usepackage[frozencache,cachedir=.]{minted}
\usepackage{enumitem,amssymb}
\usepackage{pifont}

\floatstyle{plain} 
\restylefloat{listing}

\title{\LARGE \bf

\vspace{1cm}

Taming the Panda with Python: A Powerful Duo for Seamless Robotics Programming and Integration

\vspace{-10pt}
}

\author{\parbox{6.8 in} {\centering Jean Elsner$^{1,*}$
}
\thanks{$^{1}$ Munich Institute of Robotics and Machine Intelligence (MIRMI), Technical University of Munich (TUM), Munich, Germany}
\thanks{$*$ Correspondence: jean.elsner@tum.de}
}

\begin{document}
\definecolor{bg}{rgb}{0.95,0.95,0.95}
\markboth{Preprint Version.}
{Elsner: Taming the Panda with Python} 

\maketitle
\begin{abstract}
Franka Emika robots have gained significant popularity in research and education due to their exceptional versatility and advanced capabilities. This work introduces panda-py -- a Python interface and framework designed to empower Franka Emika robotics with accessible and efficient programming. The panda-py interface enhances the usability of Franka Emika robots, enabling researchers and educators to interact with them more effectively. By leveraging Python's simplicity and readability, users can quickly grasp the necessary programming concepts for robot control and manipulation. Moreover, integrating panda-py with other widely used Python packages in domains such as computer vision and machine learning amplifies the robot's capabilities. Researchers can seamlessly leverage the vast ecosystem of Python libraries, thereby enabling advanced perception, decision-making, and control functionalities. This compatibility facilitates the efficient development of sophisticated robotic applications, integrating state-of-the-art techniques from diverse domains without the added complexity of ROS.
\end{abstract}
\IEEEpeerreviewmaketitle
\section{Introduction}
In recent years, Python has emerged as a dominant language in the machine learning community, thanks to its extensive libraries and frameworks such as TensorFlow, PyTorch, and scikit-learn~\cite{tensorflow,pytorch,sklearn_api}. However, its popularity is not limited to machine learning alone. Python is gaining significant traction in the robotics community as well~\cite{rtb}. While there have been occasional voices of concern from the robotics community regarding Python's performance for real-time and resource-intensive robotics tasks, it is worth noting that performance-critical components can be implemented in languages like C or C++ and seamlessly integrated with Python~\cite{boost-python}. This combination allows developers to harness the high-level features and ease of use provided by Python while still achieving the desired performance~\cite{pybind11}. Additionally, Python's cross-platform compatibility, portability, and extensive ecosystem of libraries make it an attractive choice for robotics. The language's ease of use and rapid prototyping capabilities further contribute to its growing adoption in the robotics community, enabling researchers and developers to quickly iterate, experiment, and deploy robotics systems with ease. Finally, the programming language's immense popularity and ease of use make it an excellent choice for robotics education, enabling students to quickly learn and experiment with robotics concepts.

Franka Emika robot manipulators have gained significant popularity in research and education due to their exceptional capabilities and versatility. These robots are highly sought after for their industrial repeatability, force sensitivity, and torque control interface, making them well-suited for various applications. The ability to precisely repeat tasks with high accuracy and their sensitive force feedback capabilities, enables researchers to explore areas such as human-robot interaction, collaborative robotics, and intricate manipulation tasks~\cite{franka_emika_robot}. Efforts are already underway to integrate Franka Emika robots into educational settings, including schools in Germany, where the user-friendly browser-based interface called "Desk" enables users to program the robots graphically using drag and drop for simple tasks, facilitating robotics education at various levels~\cite{roboterfabrik}. However, for more advanced applications, users will have to use either the provided interface through an open-source C++ library called libfranka directly or a ROS version of the same interface. The C++ library has rather strict real-time requirements, and setting it up and effectively programming it can be a daunting task for novice programmers. Similarly, using ROS comes with a considerable overhead, a steep learning curve, and limited portability and cross-platform support.

The panda-py\footnote{Earlier models of the Franka Emika robot were known as Panda, hence the name.} framework simplifies the programming, deployment, and installation process for Franka Emika robot systems. It offers an all-in-one solution with pre-packaged dependencies, ensuring a seamless experience out of the box. With panda-py, researchers and developers can focus on their work without the hassle of manual setup, benefiting from its user-friendly interface, extensive Python ecosystem, and effortless integration. It streamlines the process, making programming and experimentation with Franka Emika robots more accessible and efficient for research and education purposes. In this paper, the author aims to demonstrate the utility of panda-py in a tutorial style\footnote{All of the provided examples are also available online and are ready to run on real hardware.}.

\section{Installation and Setup}
The panda-py software is implemented as a Python package. Specifically, it is distributed as a Python wheel, i.e., a pre-built binary that can be installed using the package manager pip. The wheel includes all the needed dependencies to connect to and control the robot, while the package manager will install the appropriate versions for the local platform. To install the package, execute

\begin{minted}[mathescape,
               gobble=2,
               bgcolor=bg,
               framesep=2mm]{bash}
  pip install panda-python
\end{minted}
from a terminal\footnote{Visit the online repository at \url{http://github.com/JeanElsner/panda-py} for more information on how to build from source or install specific versions.}. Upon initiating the robotic system, a prerequisite for controlling the robot involves the release of its brakes and the activation of the interface. Typically, this procedure is done through the browser-based Desk interface~\cite{franka_emika_robot}. Nonetheless, this approach may prove inconvenient when dealing with headless setups or highly integrated systems. To address this challenge, panda-py incorporates a specialized Desk client, facilitating the seamless execution of these essential tasks more efficiently and user-friendly. (cf. Code Block~\ref{code:unlock}).

\newminted{python}{mathescape,
               linenos,
               numbersep=2pt,
               gobble=2,
               bgcolor=bg,
               breaklines=true,
               framesep=2mm}

\begin{listing}[H]
\begin{pythoncode}
  import panda_py
  
  desk = panda_py.Desk(hostname, username, password)
  desk.unlock()
  desk.activate_fci()
\end{pythoncode}
\caption{Use the Desk client to connect to the web application on the control unit to unlock the brakes and activate the Franka Research Interface (FCI) for robot torque control.}
\label{code:unlock}
\end{listing}
Once the robot is prepared, a connection can be established by instantiating the Panda class with the robot's hostname as an argument, as exemplified in Code Block~\ref{code:connect}.

\begin{listing}[H]
\begin{pythoncode}
  from panda_py import libfranka
  
  panda = panda_py.Panda(hostname)
  gripper = libfranka.Gripper(hostname)
\end{pythoncode}
\caption{Connect to the robot using the Panda class. The default gripper from Franka Emika does not support real-time control and can be controlled using the libfranka bindings directly.}
\label{code:connect}
\end{listing}
The Panda class is a high-level wrapper with various convenience functions over libfranka's robot class. However, panda-py also includes bindings for all the low-level libfranka types and functions as part of a subpackage that may be used directly. This feature is used in Code Block~\ref{code:connect} to connect to the Franka Emika Hand.

For the remainder of the tutorial, we will assume that a connection to the robot hardware was established, and the Panda and Gripper instances were assigned to the variables panda and gripper, respectively.

\section{Basic Robot Control}
The panda-py package offers its users powerful features out of the box. There are modules for time-optimal motion generation~\cite{motion}, analytical inverse kinematics~\cite{ik}, a library of proven robust standard controllers, integrated state logging, and more. These components are implemented as CPython modules in C++ and can seamlessly be used in Python code. Table \ref{tab:runtimes} compares the runtimes of common API calls between the Python bindings and native C++. Motion generation can be accessed through the methods of the Panda class. The robot's neutral or starting pose can be reached with a single call to move\_to\_start. The phase space around this pose is characterized by high manipulability, reachability, and distance to joint limits. Code Block~\ref{code:joint-motion} further demonstrates motion in joint space by adding a displacement to the homogeneous transform describing the end-effector pose and using the built-in inverse kinematics to compute the goal joint positions.

\setlength{\tabcolsep}{10pt}
\renewcommand{\arraystretch}{1.5}
\begin{table}[h]
\centering
\begin{tabular}{|l|l|l|}
\hline
API Call        & Python Runtime (\si{\second}) & C++ Runtime (s) \\ \hline
fk              & \num{2.49e-06} & \num{1.84e-06}    \\ 
ik              & \num{2.34e-06} & \num{7.65e-07}    \\ 
JointMotion     & \num{1.73e-02} & \num{1.70e-02}    \\ 
CartesianMotion & \num{9.66e-03} & \num{8.10e-03}    \\ \hline
\end{tabular}
\caption{Average runtimes of common panda-py API calls executed in Python 3.10 and C++17. Each call was executed and timed \num{10000} times with random state-space samples (5 waypoints for motion generation) on an Intel\textregistered~Core\texttrademark~i7-8565U CPU @ 1.80GHz.}
\label{tab:runtimes}
\end{table}

\begin{listing}[H]
\begin{pythoncode}
  panda.move_to_start()
  pose = panda.get_pose()
  pose[2,3] -= .1
  q = panda_py.ik(pose)
  panda.move_to_joint_position(q)
\end{pythoncode}
\caption{Simple motion generation in joint space. The call to get\_pose produces a $4\times 4$ matrix representing the homogeneous transform from robot base to end-effector. The indices $2,3$ refer to the third row and fourth column, respectively, i.e., the z-coordinate. The position in z is lowered by $\SI{0.1}{\meter}$ and passed to the inverse kinematics function to produce joint positions. Finally, the call to move\_to\_joint\_position generates a motion from the current to the desired joint potions.}
\label{code:joint-motion}
\end{listing}

Similarly, Cartesian motions can be executed directly by providing a goal pose as seen in Code Block~\ref{code:cart-motion}. Note that the interface can also compute motions with multiple waypoints. The integrated motion generators will compute time-optimal trajectories between the points while ensuring the robot adheres to velocity, acceleration, and joint limit constraints. Internally the trajectory will be traced by a joint impedance controller. In panda-py, the user can set various advanced parameters, such as control gains and allowed path deviation, when generating motion.

\begin{listing}[H]
\begin{pythoncode}
  panda.move_to_start()
  pose = panda.get_pose()
  pose[2,3] -= .1
  panda.move_to_pose(pose)
\end{pythoncode}
\caption{Simple motion generation in Cartesian space. The z-coordinate of the current end-effector pose is lowered by $\SI{0.1}{\meter}$ as in Code Block \ref{code:joint-motion}. However, the resulting pose is passed directly to move\_to\_pose to produce a motion in Cartesian space.}
\label{code:cart-motion}
\end{listing}

While the software features numerous shorthands and convenience methods, users can always access the full breadth of the libfranka library. For instance, a call to

\begin{minted}[mathescape,
               gobble=2,
               bgcolor=bg,
               framesep=2mm]{bash}
  panda.get_state()
\end{minted}

will retrieve the latest libfranka RobotState received from the robot. Similarly, the Panda class also provides a reference to the libfranka Model associated with the running instance.

\begin{minted}[mathescape,
               gobble=2,
               bgcolor=bg,
               framesep=2mm]{bash}
  panda.get_model()
\end{minted}

These Python wrappers offer the entire C++ API, i.e., class member variables and functions. The same is true for the previously initialized gripper. The gripper can be controlled using the class member functions grasp and move.

\begin{minted}[mathescape,
               gobble=2,
               bgcolor=bg,
               breaklines=true,
               framesep=2mm]{bash}
  gripper.grasp(width, speed, force, epsilon_inner, epsilon_outer) 
\end{minted}

\begin{minted}[mathescape,
               gobble=2,
               bgcolor=bg,
               breaklines=true,
               framesep=2mm]{bash}
  gripper.move(width, speed) 
\end{minted}

Function calls in panda-py allow users to use native Python types as arguments. More than that, the backend uses the powerful Eigen~\cite{eigen} library for linear algebra and will transparently and efficiently convert Eigen matrices to numpy~\cite{numpy} arrays without modifying the underlying memory structure.

\begin{figure}[h]
    \centering
    \includegraphics[width=0.5\textwidth]{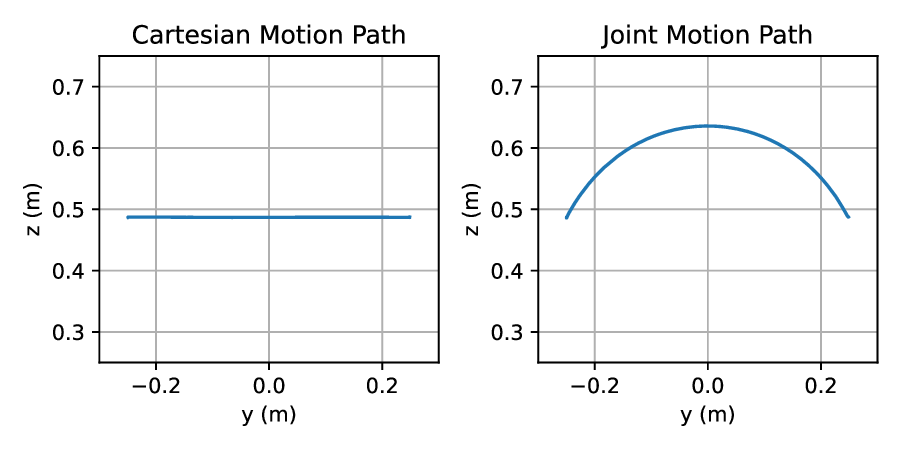}
    \caption{Plotting the robot's internal state during motion recorded with the integrated logging mechanism. The plots show the traced path of the end-effector in the robot base frame projected onto the xy-plane. The motion is generated as seen in Code Block~\ref{code:logging}, i.e., using two waypoints displaced \SI{0.5}{\meter} along the y-axis. The motion of the left plot was generated using move\_to\_pose and resulted in a linear path in Cartesian space. In contrast, the motion of the right plot is the result of a call to move\_to\_joint\_position with the same waypoints, resulting in a non-linear path in Cartesian space.}
    \label{fig:plot-from-log}
\end{figure}

Logging robot states is a ubiquitous requirement in robotics experiments, yet it can be challenging to set up, particularly when capturing states at high frequencies. However, panda-py offers a convenient solution with its integrated mechanism to write the whole state of the robot into a circular buffer at \SI{1}{\kilo\hertz} when activated. This feature simplifies the logging process, allowing users to easily capture and store data for subsequent evaluation, plotting, and other signal-processing tasks. Code Block~\ref{code:logging} illustrates how to enable logging and store the resulting buffer, while Figure~\ref{fig:plot-from-log} shows plots of some of the captured telemetry.

\begin{listing}[H]
\begin{pythoncode}
  from panda_py import constants

  T_0 = panda_py.fk( constants.JOINT_POSITION_START)
  T_0[1,3] = 0.25
  T_1 = T_0.copy()
  T_1[1,3] = -0.25

  panda.move_to_pose(T_0)
  panda.enable_logging(2e3)
  panda.move_to_pose(T_1)
  panda.disable_logging()
  log = panda.get_log()
\end{pythoncode}
\caption{Using the integrated logging mechanism, the libfranka RobotState can be logged at a frequency of \SI{1}{\kilo\hertz}. Based on the starting pose, this example creates two end-effector poses, T\_0 and T\_1, displaced \SI{0.25}{\meter} to the left and right, respectively. Logging is enabled for this Panda instance before a motion is generated between these poses (line 9). The enable\_logging function takes the buffer size in the number of steps as an argument. As such, \num{2e3} steps at \SI{1}{\kilo\hertz} correspond to a buffer holding the state of the past 2 seconds. After the motion is finished, logging is disabled, and the buffer is retrieved (line 12).}
\label{code:logging}
\end{listing}

Finally, for more advanced applications, there is a library of standard controllers. Controllers are classes instantiated by users and passed to the Panda class for execution. The controllers will run independently of the Python Global Interpreter Lock in the background and meet all of libfranka's real-time requirements. At the same time, the user may provide an input signal asynchronously.

\begin{listing}[H]
\begin{pythoncode}
  import numpy as np

  panda.move_to_start()
  ctrl = controllers.CartesianImpedance()
  x0 = panda.get_position()
  q0 = panda.get_orientation()
  runtime = np.pi*4.0
  panda.start_controller(ctrl)

  with panda.create_context(frequency=1e3, max_runtime=runtime) as ctx:
    while ctx.ok():
      x_d = x0.copy()
      x_d[1] += 0.1*np.sin(ctrl.get_time())
      ctrl.set_control(x_d, q0)
\end{pythoncode}
\caption{Running a panda-py controller. After initializing the controller, the current position and orientation are stored in x0 and q0, respectively, where q0 is a quaternion representation of the end-effector orientation. After starting the controller, a PandaContext is created from the Panda object (line 10). PandaContext is a convenient context manager that executes a loop at a fixed frequency. The call to PandaContext.ok throttles the loop and raises any control exceptions that may have been raised by libfranka. The use of PandaContext is optional, and users are free to manage the control flow how they wish. This example adds a periodic linear displacement along the y-axis to the initial pose (line 13). This results in the end-effector moving periodically from left to right in straight lines.}
\label{code:controller}
\end{listing}

Code Block~\ref{code:controller} demonstrates the essentials of using a panda-py controller. The controller CartesianImpedance is instantiated and passed to the Panda class for execution. While the controller is running, the user can use its set\_control method asynchronously to provide an input signal. The panda-py controllers provide many configuration options specific to the individual controller, such as control gains or filter settings. These options may be changed at run-time as well. In this example, the input signal is provided in a high-frequency loop without filtering for more fine-grained control. All controllers include virtual joint walls, so the input signal need not explicitly consider the joint limits. While various methods exist to control aspects such as the robot's reflex behavior and error recovery, the content covered in this chapter already provides a foundational level of control over the robot.

\section{Integration with Python Packages}

A notable aspect of panda-py is its ability to seamlessly integrate with popular Python packages, such as the Robotics Toolbox for Python~\cite{rtb} or MuJoCo~\cite{mujoco}. Using Python bindings to integrate the robot hardware makes for singularly lightweight and straightforward integration. The middleware layer can be avoided entirely while all necessary hardware setup and preparation can be centralized. In our previous work, we found this aspect particularly convenient when running MuJoCo simulations with hardware in the loop for haptic interaction and predictive modeling~\cite{parti}.

By leveraging the capabilities of these existing packages, researchers can easily extend the functionalities of Franka Emika robots, as is demonstrated in this final example. Specifically, a resolved rate motion controller utilizing reactive control based on quadratic programming is integrated with panda-py, leveraging the Robotics Toolbox for Python. This integration exemplifies the ease of extending panda-py to incorporate more complex functionalities, such as reactive collision avoidance and mobile manipulation. The code snippet in Code Block~\ref{code:rtb} provides a clear representation of the integrated implementation. Notably, the solution to the controller's quadratic program yields joint velocities that seamlessly interface with the IntegratedVelocity controller.

\begin{pythoncode}
  import qpsolvers as qp
  import roboticstoolbox as rtb
  import spatialmath as sm
  
  ctrl = controllers.IntegratedVelocity()
  panda.move_to_start()
  panda.start_controller(ctrl)

  # Initialize roboticstoolbox model
  panda_rtb = rtb.models.Panda()

  # Set the desired end-effector pose
  Tep = panda_rtb.fkine(panda.q) * sm.SE3(0.3, 0.2, 0.3)

  # Number of joints in the panda which we are controlling
  n = 7
  arrived = False

  # The original example runs in simulation with a control frequency of 20Hz
  with panda.create_context(frequency=20) as ctx:
    while ctx.ok() and not arrived:

      # The pose of the Panda's end-effector
      Te = panda_rtb.fkine(panda.q)

      # Transform from the end-effector to desired pose
      eTep = Te.inv() * Tep

      # Spatial error
      e = np.sum(np.abs(np.r_[eTep.t, eTep.rpy() * np.pi / 180]))

      # Calulate the required end-effector spatial velocity for the robot
      # to approach the goal. Gain is set to 1.0
      v, arrived = rtb.p_servo(Te, Tep, 1.0)

      # Gain term (lambda) for control minimisation
      Y = 0.01

      # Quadratic component of objective function
      Q = np.eye(n + 6)

      # Joint velocity component of Q
      Q[:n, :n] *= Y

      # Slack component of Q
      Q[n:, n:] = (1 / e) * np.eye(6)

      # The equality contraints
      Aeq = np.c_[panda_rtb.jacobe(panda.q), np.eye(6)]
      beq = v.reshape((6,))

      # Linear component of objective function: the manipulability Jacobian
      c = np.r_[ -panda_rtb.jacobm().reshape((n,)), np.zeros(6)]

      # The lower and upper bounds on the joint velocity and slack variable
      lb = -np.r_[panda_rtb.qdlim[:n], 10 * np.ones(6)]
      ub = np.r_[panda_rtb.qdlim[:n], 10 * np.ones(6)]

      # Solve for the joint velocities dq
      qd = qp.solve_qp(Q, c, None, None, Aeq, beq, lb=lb, ub=ub, solver='daqp')

      # Apply the joint velocities to the Panda
      ctrl.set_control(qd[:n])

\end{pythoncode}
\captionof{listing}{Resolved rate controller with reactive manipulability maximization. This example is from the Robotics Toolbox for Python~\cite{mmc}. To run it on the real hardware with panda-py requires only connecting the inputs and outputs of the control loop to panda-py, i.e., using the joint positions Panda.q and providing the control signal to IntegrateVelocity.set\_control. Additionally, the inequality constraints to avoid the joint limits were removed, as panda-py controllers already have integrated joint limit avoidance using impedance control.}
\label{code:rtb}

\section{Conclusion}
In conclusion, this research paper has introduced panda-py as a Python interface and framework that facilitates the programming of Franka Emika robots. The inclusion of concise and approachable code examples throughout the paper highlights panda-py's user-friendly nature and effectiveness in controlling the robots. It is worth noting that this paper provides only a glimpse into the extensive capabilities of panda-py, as it represents a dynamic and evolving ecosystem. Online resources, including additional examples, documentation, and the help of the robotics community, contribute to the continual maintenance and expansion of panda-py. Researchers and users are encouraged to explore these resources for a comprehensive understanding of panda-py's potential. The author aims to apply the same paradigm and lessons learned from developing panda-py to other robot hardware in future work. Additionally, integrating panda-py with reinforcement learning environments opens up exciting opportunities to explore robot learning.

\section*{ACKNOWLEDGMENT}
The author gratefully acknowledges the funding support provided by the Lighthouse Initiative Geriatronics by StMWi Bayern (Project X, grant no. IUK-1807-0007// IUK582/001).

\bibliographystyle{IEEEtran}
\tiny{
\bibliography{references}}
\end{document}